\documentclass{article}
\usepackage[english]{babel}
\usepackage[letterpaper,top=2cm,bottom=2cm,left=3cm,right=3cm,marginparwidth=1.75cm]{geometry}
\usepackage{amsmath,amsfonts}
\usepackage{algorithmic}
\usepackage{algorithm}
\usepackage{array}
\usepackage[caption=false,font=normalsize,labelfont=sf,textfont=sf]{subfig}
\usepackage{textcomp}
\usepackage{stfloats}
\usepackage{url}
\usepackage{verbatim}
\usepackage{graphicx}
\usepackage{cite}
\usepackage{multirow}
\usepackage{authblk}
\usepackage{hyperref}

\providecommand{\keywords}[1]
{
  \small	
  \textbf{\textit{Keywords---}} #1
}

\title{Deep Learning as Ricci Flow}

\author[1,2]{Anthony Baptista \thanks{abaptista@turing.ac.uk}}
\author[1]{Alessandro Barp}
\author[1]{Tapabrata Chakraborti}
\author[3]{Chris Harbron}
\author[1,4,5]{ Ben D. MacArthur}
\author[1,6]{Christopher R. S. Banerji \thanks{cbanerji@turing.ac.uk}}

\affil[1]{The Alan Turing Institute, The British Library, London, NW1 2DB, United Kingdom}
\affil[2]{School of Mathematical Sciences, Queen Mary University of London, London, E1 4NS, United Kingdom}
\affil[3]{Roche Pharmaceuticals, Welwyn Garden City, AL7 1TW, United Kingdom}
\affil[4]{School of Mathematical Sciences, University of Southampton, Southampton SO17 1BJ, United Kingdom}
\affil[5]{Faculty of Medicine, University of Southampton, Southampton SO17 1BJ, United Kingdom}
\affil[6]{University College London Hospitals, NHS Foundation Trust, London, NW1 2BU, United Kingdom}

\begin{document}

\maketitle

\begin{abstract}
Deep neural networks (DNNs) are powerful tools for approximating the distribution of complex data. It is known that data passing through a trained DNN classifier undergoes a series of geometric and topological simplifications. While some progress has been made toward understanding these transformations in neural networks with smooth activation functions, an understanding in the more general setting of non-smooth activation functions, such as the rectified linear unit (ReLU), which tend to perform better, is required.
Here we propose that the geometric transformations performed by DNNs during classification tasks have parallels to those expected under Hamilton's Ricci flow - a tool from differential geometry 
that evolves a manifold by smoothing its curvature, in order to identify its topology. To illustrate this idea, we present a computational framework to quantify the geometric changes that occur as data passes through successive layers of a DNN, and use this framework to motivate a notion of `global Ricci network flow' that can be used to assess a DNN's ability to disentangle complex data geometries to solve classification problems. By training more than $1,500$ DNN classifiers of different widths and depths on synthetic and real-world data, we show that the strength of global Ricci network flow-like behaviour correlates with accuracy for well-trained DNNs, independently of depth, width and data set. Our findings motivate the use of tools from differential and discrete geometry to the problem of explainability in deep learning.
\end{abstract}

\section{Introduction}

Artificial neural networks are ubiquitous tools in applied science, with remarkable success in understanding the distribution of complex data, and relationships between features \cite{kotsiantis2007supervised}. This power is largely due to their capacity to closely approximate unknown, non-linear, multi-dimensional functions \cite{LeCun2015}. Classical work on universal approximation demonstrated that given sufficient width, a neural network with a single hidden layer can approximate any Borel measurable function \cite{cybenko1989approximation,hornik1989multilayer}. Subsequently, deep neural networks (DNNs) comprising series of linearly weighted connections between multiple hidden layers of non-linear activation functions \cite{LeCun2015}, were demonstrated to be exponentially more efficient (in terms of the number of hidden nodes required) at approximating certain classes of function, compared to their single layer counterparts \cite{montufar2014number,Delalleau11,mhaskar2016learning}. Informally, each layer of a DNN provides a compositional representation of the function being approximated, which the next layer can build upon in a hierarchical way \cite{pascanu2013number,mhaskar2016deep}. This step-wise approach is intuitively more efficient than a single layer neural network and empirical results have accordingly demonstrated the superiority of these deeper architectures \cite{krizhevsky2012imagenet,goodfellow2013multi}. 

Despite this intuition, theoretical understanding of precisely how DNNs achieve such remarkable results lags behind their practical application \cite{serra2018bounding}, although this issue has been approached from numerous perspectives. For instance, theoretical results on hyperplane arrangement have provided upper bounds on the number of linear `pieces' that a network's output function can describe, when the activation functions at hidden nodes are all piece-wise linear \cite{montufar2014number,pascanu2013number}. This reasoning revealed that deep networks have the capacity to readily construct more complex outputs (i.e., with more linear `pieces') than shallow networks with the same number of nodes.
These theoretical results notwithstanding, empirical studies have demonstrated that under the practical constraints of training by gradient descent, the number of linear regions constructed by deep and shallow network output functions is often comparable \cite{hanin2019deep}. Approaches which combine theory with the practical constraints of DNN training are clearly needed to better understand their success.

A complementary strand of theoretical reasoning considers how input data is sequentially deformed as it passes through successive layers of a DNN \cite{Brahma2016,Naitzat2020,cohen2020separability,ansuini2019intrinsic,wheeler2021activation}. Employing techniques for studying the discrete geometry and topology of data clouds \cite{magai2022topology,Naitzat2020} it has been proposed that data drawn from complex, non-linearly separable classes, are sequentially deformed by trained DNN classifiers into linearly separable structures at the output layer \cite{cohen2020separability}. In support of this hypothesis, studies using tools from computational topology have shown that data sampled from a pair of entangled manifolds undergoes topological and geometric simplifications as it passes through the layers of an optimised feed-forward DNN binary classifier - ultimately mapping in the final layer to two distinct hyper-spheres \cite{Naitzat2020}. Subsequent studies have associated the ability of a DNN to perform such simplifications with higher classification accuracy \cite{magai2022topology,wheeler2021activation}. Importantly, it was observed that these topological simplifications do not always occur monotonically as data passes through successive layers of a DNN \cite{wheeler2021activation}, suggesting that this disentanglement is not a straightforward gradient process. 

Investigations of intrinsic dimension (the minimum number of dimensions required to describe a data set), by mean field theory \cite{cohen2020separability} and nearest neighbour approaches \cite{ansuini2019intrinsic}, have similarly found that data is geometrically `simplified' (i.e., its intrinsic dimension is reduced) as it passes through the layers of a DNN. While data representations at successive DNN layers will typically retain a non-linear character \cite{ansuini2019intrinsic}, a popular heuristic is that the total curvature of the data representation reduces as data passes through successive layers, and though some empirical evidence supports such data `flattening' \cite{Brahma2016}, it is not ubiquitous \cite{ansuini2019intrinsic}.

The notion that DNNs orchestrate topological and geometric changes in data structures has motivated some to consider differential geometry approaches to the analysis of DNN efficacy \cite{hauser2017principles}. For example, a Riemannian geometry framework has been proposed in which each activation layer of the DNN is considered as a coordinate transformation of the previous layer, finally resulting in a coordinate system in which input data classes are linearly separable \cite{hauser2017principles}. While elegant, this analytical formulation relies on continuity assumptions and so does not directly apply to learning with non-smooth activation functions, such as the widely used rectified linear unit (ReLU) \cite{BENFENATI2023331}, and its variations. A similar line of research interprets the transformations performed by DNNs from a dynamical systems perspective~\cite{haber2017stable,lu2018beyond,sander2022sinkformers}. While this framework can be used to design and improve the stability of DNN architectures, it also makes continuity assumptions that are not satisfied by widely used models that employ non-smooth activation functions. New geometric tools, that facilitate practical assessment of the topological simplifications of deep learning classifiers are thus needed. 

Here, we outline a new approach that makes use of the fact that DNNs trained for classification tasks tend to transform a data cloud until each class is linearly separable.
Underpinning this observation is a `manifold hypothesis' in three parts: (1) given a set of data points sampled from a (possibly high dimensional) continuous feature space, we expect that associations between the features will force the sampled points to lie on a (possibly disconnected) subset, or manifold, of the feature space. (2) For a separable classification problem, the challenge facing the DNN is then to disentanlge important topological properties of the manifold from its, perhaps very complex, geometry. (3) The DNN solves this problem by evolving the manifold such that its geometry is unravelled, to reveal topological properties of the data relevant to the problem at hand. Viewed from this perspective, the dynamics associated with the DNN can therefore be seen as an example of a geometric `flow'.

A well known tool in differential geometry for understanding evolving manifolds, with a rich theoretical and empirical literature, is Ricci flow - a partial differential equation that was introduced by Hamilton to evolve a metric on a compact manifold according to its Ricci curvature \cite{HamiltonR.1988}, and was famously central to the  proof of the Poincar{\' e} conjecture by Perelman \cite{Perelman2003,Perelman2002}. Intuitively, Ricci flow smooths irregularities in a complex manifold, and allows us to obtain topological information from its geometry, by evolving the initial complex manifold towards its connected sum decomposition into simpler (locally homogeneous) manifolds~\cite{anderson2004geometrization}.
This has appealing parallels with the geometric view of deep learning, articulated above. However, while Ricci flow is a time continuous process, the geometric view of deep learning is discrete in both time (due to the fact that learning occurs via a sequence of discrete transformations, associated with the network layers) and space (due to the fact that the training data constitutes a point sample from the underlying manifold, not the manifold per se). 

Importantly, discrete analogues of Ricci curvature have been developed from consideration of the structure of simplical complexes \cite{Forman2003} and optimal transport processes \cite{Ollivier2007} which allow construction of discrete Ricci flows, with numerous applications in network theory \cite{weber2016,weber2017,Cohen2022,Ni2019,topping2021understanding}.

In machine learning contexts, metric flows, of which Ricci flow is an example, have been used to interpret the training process of DNNs~\cite{halverson2023metric,gukov2024rigor}. Discrete Ricci flow has been used for community detection in weighted networks - a problem with close parallels to classification \cite{Ni2019} - and has proven useful for rewiring graph neural networks to avoid bottlenecks \cite{topping2021understanding}. Crucially, these applications show that discrete Ricci flow shares with its continuous analogue the capacity to evolve the geometry of a network (here, a discrete analogue of a manifold) by stretching regions of negative curvature. 

Here, we construct an empirical framework to quantify Ricci flow-like behaviour in a well-trained binary classifier DNN. By performing experiments on a range of synthetic and real data sets, using DNNs with a variety of architectures, we find that Ricci flow-like behaviour can generally be seen, so long as the data and its representations at DNN layers are observed at an appropriate scale. Moreover, we observe that stronger Ricci flow-like behaviour is positively associated with classification accuracy, independently of DNN width, depth and data set. Based on these results we propose that Ricci flow metrics can be used to optimise DNN architecture and possibly inform data point specific uncertainty estimates. Collectively, these findings motivate further study of how tools from differential geometry can be used to aid DNN design and improve our understanding of the theoretical basis of deep learning. 

\section{Methods}

\subsection{Binary classification with deep learning}

Following the notation of \cite{Naitzat2020} we consider a binary classification problem: given a set of randomly sampled labelled data points $X \times Y \subset M \times \{a,b\}$, where $M \subset \mathbb{R}^N$ is a compact manifold given by the union of submanifolds $M=M_{a} \cup M_{b}$,  for each $x \in X$ we wish to develop a procedure to correctly determine whether it was drawn from $M_a$ or $M_b$.

We propose to solve this problem with a feed-forward DNN containing $L$ hidden layers. Following standard construction, at each layer $l=1,...,L$, comprising $N_l$ nodes, we generate an output $x_l$ via:
\begin{align*}
    h_l &= W_l x_{l-1} + b_l\\
    x_l &= \psi(h_l)
\end{align*}
where $x_0 \in X$, $W_l\in \mathbb{R}^{N_{l-1} \times N_{l}}$ is a matrix describing weights on edges connecting nodes in layer $l-1$ to layer $l$, $b_l\in \mathbb{R}^{N_l}$ is a vector of biases and $\psi:\mathbb{R}^{N_{l}}\rightarrow\mathbb{R}^{N_{l}}$ is a non-linear activation function implemented by each node. Here we choose the ReLu operator:
\begin{equation}
    \psi(x) = \max(x,0),
\end{equation}
which was previously found to be the most efficient function for topological simplification of data through the layers of a DNN \cite{Naitzat2020}. The activation function for the output layer is chosen as the sigmoid function.

The weights $W_l$ and biases $b_l$ are optimised using a standard RMSprop optimiser, with learning rate 0.001 and a binary cross-entropy loss function, using a labelled training data set $X_{\textrm{train}} \subset X$, employing \emph{keras} (v. 2.9) \cite{chollet2015keras}.

The trained DNN can then be applied to an unseen test data set $X_{\textrm{test}} \subset X/X_{\textrm{train}}$, for bench-marking and further analysis. Here we will focus on how $X_{\textrm{test}}$ is deformed by each layer using geometric tools associated with discrete Ricci flow. 

\subsection{Discrete Ricci flow}

For a compact manifold $M$ a Ricci flow evolves a Riemannian metric $g_t$ (which describes  infinitesimal distances)  such that:
\begin{equation} \label{RF}
    \frac{\partial g_t}{\partial t} = -\alpha \textrm{Ric}(g_{t})
\end{equation}
where $\textrm{Ric}(g_t)$ is the Ricci curvature of $M$ at time $t \in (a,b)$ and $\alpha\in \mathbb{R}^{+}$. 
The choice $\alpha=2$ is conventional but any positive real-valued scalar will guarantee existence of a solution in finite time. 
The negative sign ensures the equation is parabolic and thus the dynamics tend to  `average out' the curvature, just like heat flow averages out the distribution of heat.
% Integrating Eq.~\ref{RF} using Euler's forward method, with step size $\Delta t$, gives the following approximation:
% \begin{equation} \label{fE}
%     g_{t+ \Delta t} - g_{t} \approx  - \alpha \Delta t \, \textrm{Ric}(g_t).
% \end{equation}

To investigate whether Ricci flow-like behaviour is present in deep learning, we first need to adapt the above continuous formulation to a (spatially) discrete setting. 
To do so, we denote the outputs of each hidden layer $l=1,...,L$, of the trained DNN applied to the test data by $X_{\textrm{test}}^{l}$. For each layer $l=1,...,L$ of our trained DNN $X_{\textrm{test}}^{l}$ provides a sample of how representative test points are deformed by the DNN up to and including the $l$-th layer. We can obtain a coarse approximation of the geometry of this set by constructing a $k$-nearest neighbour graph $\mathcal{G}_{k}(X_{\textrm{test}}^{l}):=(V,E_{k}^{l})$, where $V$ is a set of vertices corresponding to the test data points (and thus remains fixed for each $k,l$) and $E_{k}^{l}$ is a set of edges in the $k$ nearest neighbour graph constructed on $X_{\textrm{test}}^{l}$. We approximate the distance between two points $i, j \in V$ by the shortest path distance between $i$ and $j$ in $\mathcal{G}_{k}(X_{\textrm{test}}^{l})$, which we denote $\gamma_{k}^{l}(i,j)$, 
and which forms the discrete analogue of Riemannian distance.

A number of approaches have been developed for calculating discrete Ricci curvature on graphs, including Ollivier-Ricci Curvature \cite{Ollivier2007} and Forman-Ricci Curvature \cite{Forman2003}. Here we consider the Forman-Ricci curvature, which has previously been employed for discrete Ricci flow \cite{weber2017,baptista2023charting} and which has a combinatorial construction, so is efficient to compute. 

For a graph $\mathcal{G}:=(V,E)$, with $V$ a set of vertices and $E$ a set of edges and associated vertex weights $(W_i)_{i \in V}$ and edge weights $(\omega_{ij})_{(i,j)\in E}$, the Forman-Ricci curvature $R_{\mathcal{G}}(i,j)$ on edge $(i,j) \in E$ is given by:
\begin{equation}
        R_{\mathcal{G}}(i,j) =   W_i+W_j - (\omega_{ij})^{1/2}\Bigg[W_i\sum_{k \not{=} j}(\omega_{ik})^{-1/2}+W_j\sum_{k \not{=} i}(\omega_{kj})^{-1/2}\Bigg].
\end{equation}
Whenever $(i,j) \notin E$, $R_{\mathcal{G}}(i,j)$ is not defined. 

For $\mathcal{G}_{k}(X_{\textrm{test}}^{l})$, we set edge and vertex weights to $1$ and the Forman-Ricci curvature on edge $(i,j)\in E_{k}^{l}$ simplifies to:
\begin{equation} \label{Forman}
    R_{k}^{l}(i,j) = 4- \textrm{deg}_{k}^{l}(i)- \textrm{deg}_{k}^{l}(j),
\end{equation}
where $\textrm{deg}_{k}^{l}(i)$ is the degree of vertex $i$ in $\mathcal{G}_{k}(X_{\textrm{test}}^{l})$.

\subsection{Defining global Ricci network flow-like behaviour in deep learning}

Several studies have shown that as data passes through the layers of a DNN trained to solve a classification problem, there is a geometric simplification and smoothing, which leads to a gradual separation of data points from distinct classes \cite{Naitzat2020,Brahma2016,ansuini2019intrinsic}. While it is clear that geometric and/or topological simplifications through the layers of a DNN do not necessarily proceed in a monotone manner \cite{Brahma2016,wheeler2021activation,Naitzat2020, ansuini2019intrinsic}, this behaviour does put one in mind of a Ricci flow (Figure 1). Intuitively, under a Ricci flow, regions of negative curvature in the manifold $M$ expand, while regions of positive curvature contract \cite{Ni2019}. Similarly, as data passes through a DNN we intuitively expect that, on average, data points belonging to the same class will cluster more closely together in $\mathcal{G}_{k}(X_{\textrm{test}}^{l})$, forming dense cliques with positive curvature, while disparate points will pull apart to form tree-like structures with negative curvature. In this sense, well-trained DNNs may use geometric information to evolve the discrete network representation of $X_{\textrm{\textrm{test}}}^{l}$ given by $\mathcal{G}_{k}(X_{\textrm{\textrm{test}}}^{l})$ towards a simpler architecture that is easier to cut (i.e., consisting of two dense subnetworks, sparsely connected to each other). With this in mind, it is notable that the Forman-Ricci curvature is closely related to graph Laplacian, whose second eigenpair describes the algebraic connectivity of the network and therefore the extent to which it may be easily partitioned.   %Evolving these networks via a Ricci flow would intuitively be expected to separate distinct classes.

\begin{figure} \label{fig1}
\centering
\includegraphics[width=\linewidth,keepaspectratio]{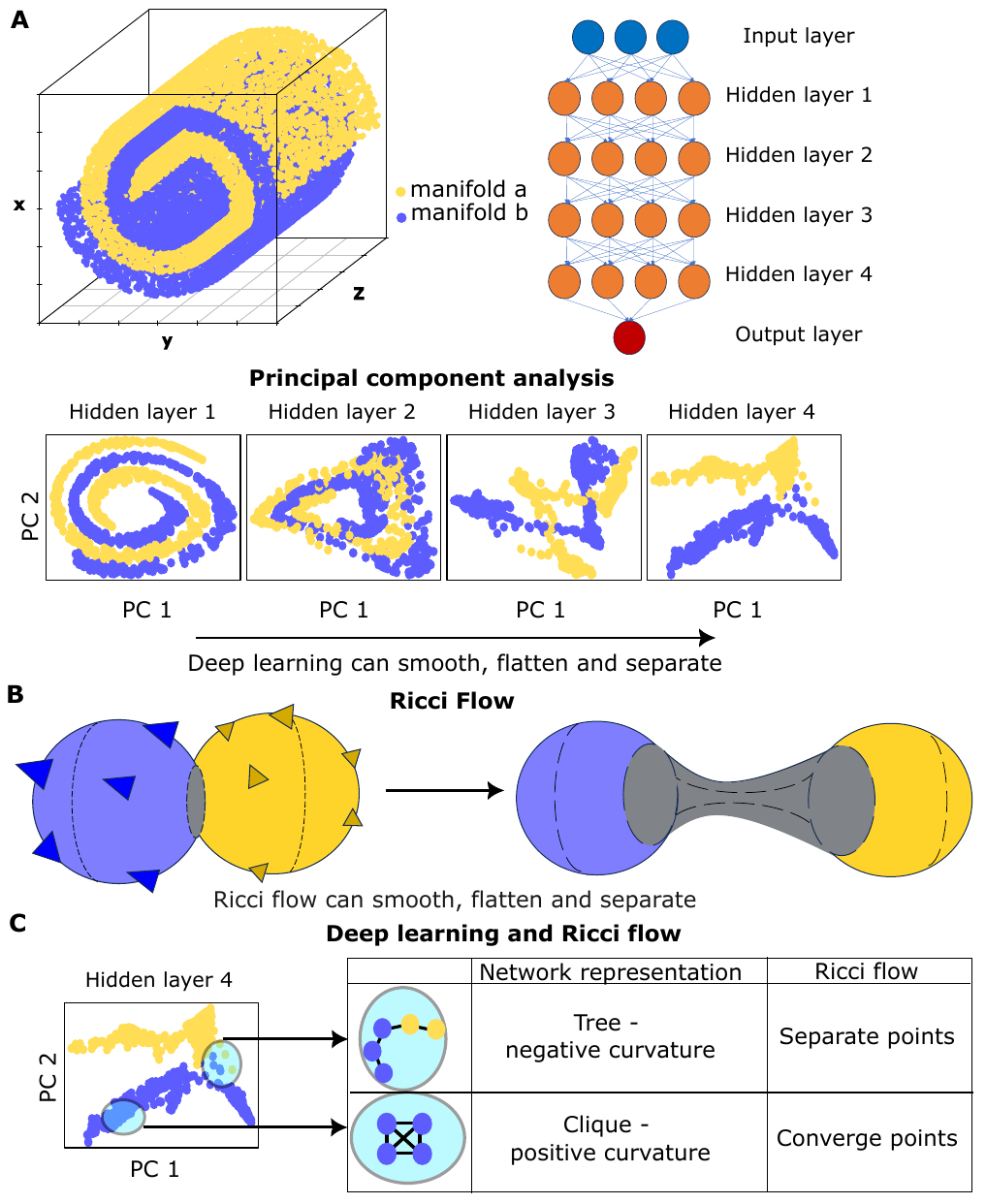}
\caption{{\it{Deep learning and Ricci flow.}} \textbf{A}. An example of deep learning. The structure of two non-linearly separable, entwined, manifolds is learned by a deep neural network (DNN). A test set, consisting of random samples drawn from the two manifolds, is passed through the trained DNN and the output of each layer is visualised via its first two principal components. As the test set passes through the layers of the trained DNN, irregularities in the geometry of the data are smoothed, and the two manifolds are separated. \textbf{B}. An example of Ricci flow. An irregular manifold, consisting of two generally positively curved regions joined by a region of negative curvature, evolves according to a Ricci flow. The irregularities on the positively curved regions are smoothed and the negatively curved region expands, separating them. \textbf{C}. When represented as a $k$-nearest neighbour graph, dense sets of points form positively curved clique-like structures which are drawn together under discrete Ricci flow; sparse sets of points form negatively curved tree-like structures, which are separated by Ricci flow.}
\end{figure}

%This intuition notwithstanding, it is clear that topological simplification through the layers of a DNN does not necessarily proceed in a monotone manner \cite{wheeler2021activation,Naitzat2020}, and that while geometric simplification of data can occur \cite{Brahma2016}, it is not guaranteed \cite{ansuini2019intrinsic}. 
%More fundamentally, we do not expect that the highly complex, non-smooth transformations characteristic of the deep learning models we consider here, to be adequately  described by a numerically integrated ordinary differential equation.

%While we do not expect geometric flows to universally describe deep learning, intuitively we do expect to observe Ricci flow-like behaviour, in the form of a series of geometric simplifications that reveal topological information, as data passes through the layers of well trained DNN. The purpose of this article is to develop a framework to observe and quantify such dynamics.

%Specifically, we expect that well-trained DNNs use geometric information to evolve the discrete network representation of $X_{\textrm{\textrm{test}}}^{l}$ given by $\mathcal{G}_{k}(X_{\textrm{\textrm{test}}}^{l})$ towards a simpler architecture that is easier to cut (i.e., consisting of two dense subnetworks, sparsely connected to each other).
%The problem of finding the best cut for the partition depends on the volume of
%the partitions~\cite{von2007tutorial}, and
%we shall use the sum of all the shortest paths as an approximation of the volume of each layer

To formalise this intuition, by analogy with Eq. \ref{RF} we therefore propose that 
\begin{equation} \label{dRF}
    g_{l+1}-g_{l} \approx -\alpha \textrm{Ric}(g_l),
\end{equation}
for each $l$, 
where
\begin{equation} 
   \textrm{Ric}(g_l) = \textrm{Ric}_l = \sum_{(i,j) \in E_{k}^{l}} R_{k}^{l}(i,j),
\end{equation}
and
\begin{equation} 
    g_l = \sum_{i,j \in V} \gamma_{k}^{l}(i,j),
\end{equation}

%These quantities are related as it can be shown there is an equivalence between shortest paths and the graph cut costs
% on planar graphs~\cite{schmidt2007efficient}.
%To interpret the change in graph volume between layers, following the Ricci flow paradigm, we shall use curvature. This choice is motivated by the relationship between the Forman-Ricci curvature and the Laplacian, whose second eigenpair describes the algebraic connectivity of the network, and the optimal solution to the graph cut problem.
%More precisely, we propose that,
% by analogy with Eq. \ref{RF},
%\begin{equation} \label{dRF}
%    g_{l+1}-g_{l} \approx -\alpha \textrm{Ric}(g_l),
%\end{equation}
%for each $l$, 
%where
%\begin{equation} 
%   \textrm{Ric}(g_l) = \textrm{Ric}_l = \sum_{(i,j) \in E_{k}^{l}} R_{k}^{l}(i,j),
%\end{equation}
whenever the data manifold passing through the layers of a well trained DNN evolves analogously to a global Ricci flow. We will refer to such dynamics global Ricci network flow.

The left hand side of Eq.~\ref{dRF} represents the total change in distance between pairs of test points at the $l$-th and $l+1$-th layers of the DNN (i.e., an estimate of the total expansion/contraction of the underlying manifold between layers $l$ and $l+1$), and the right hand side is an estimate of the total curvature of the underlying manifold at layer $l$. %We mention in passing that a parametrised family of functionals that includes the total curvature, known as Perelman $F$-functionals, can be used to understand Ricci flow as a gradient flow on the space of Riemannian metrics \cite{anderson2004geometrization}.

Denoting $\eta_l = g_{l+1} - g_l$ the sample Pearson correlation coefficient,

\begin{equation} \label{RC}
   \rho_k = \frac{\sum_{l=1}^{L-1}(\eta_l - \bar{\eta})(\textrm{Ric}_l - \bar{\textrm{Ric}})}{\sqrt{\sum_{l=1}^{L-1}(\eta_l - \bar{\eta})^2} \sqrt{\sum_{l=1}^{L-1}(\textrm{Ric}_l - \bar{\textrm{Ric}})^2}}
\end{equation}
is then a measure of the extent to which the DNN exhibits global Ricci network flow-like behaviour, where 
\begin{equation}
\bar{\eta} = \frac{1}{L-1} \sum_{l=1}^{L-1} \eta_l,
\end{equation}
and
\begin{equation}
\bar{\textrm{Ric}} = \frac{1}{L-1} \sum_{l=1}^{L-1} \textrm{Ric}_l,
\end{equation}
represent the average expansion/contraction and curvature of the manifold as it passes through successive layers of the DNN, respectively. We will refer to $\rho_k$ as the Ricci coefficient of the DNN. Negative values of the Ricci coefficient indicate the presence of global Ricci network flow-like dynamics in the DNN.

The Ricci coefficient for a DNN of depth $L$ is computed as a Pearson correlation over $L-1$ data points. To compare DNNs of various depths we therefore employed Fisher's $z$ transformation of Pearson's $r$, to calculate adjusted Ricci coefficients:
\begin{equation}
    z_k=\frac{\textrm{arctanh}(\rho_k)}{\sqrt{L-4}}.
\end{equation}

\section{Results}
\subsection{Numerical experiments}

To investigate the extent to which trained DNNs exhibit Ricci flow-like dynamics, we considered 3 synthetic data sets (Figure 2A). All describe the union of two manifolds, with varying degrees of entanglement. Data set A has been previously investigated \cite{Brahma2016} and the two manifolds, while geometrically entangled, are not topologically entangled. Data set B is similar to another previously investigated set \cite{Naitzat2020} and comprises manifolds which cannot be made linearly separable without altering their topology. Data set C comprises intersecting manifolds which are thus not-separable either by geometric or topological transformations. We also considered the well-studied MNIST data set, containing handwritten digits $0-9$ as $28 \times 28$ pixel black and white images. As simple binary classification exemplars, we considered 2 problems: first distinguishing `$1$' from `$7$' and second distinguishing `$6$' from `$8$'. Lastly we considered the fashion MNIST (fMNIST) data set, containing pictures of different items of clothing as $28 \times 28$ pixel black and white images. As simple binary classification exemplars we again considered 2 problems: first distinguishing `sandals' from `ankle boots' and second distinguishing `shirts' from `coats'.

\begin{figure}
\centering
\includegraphics[width=1\textwidth]{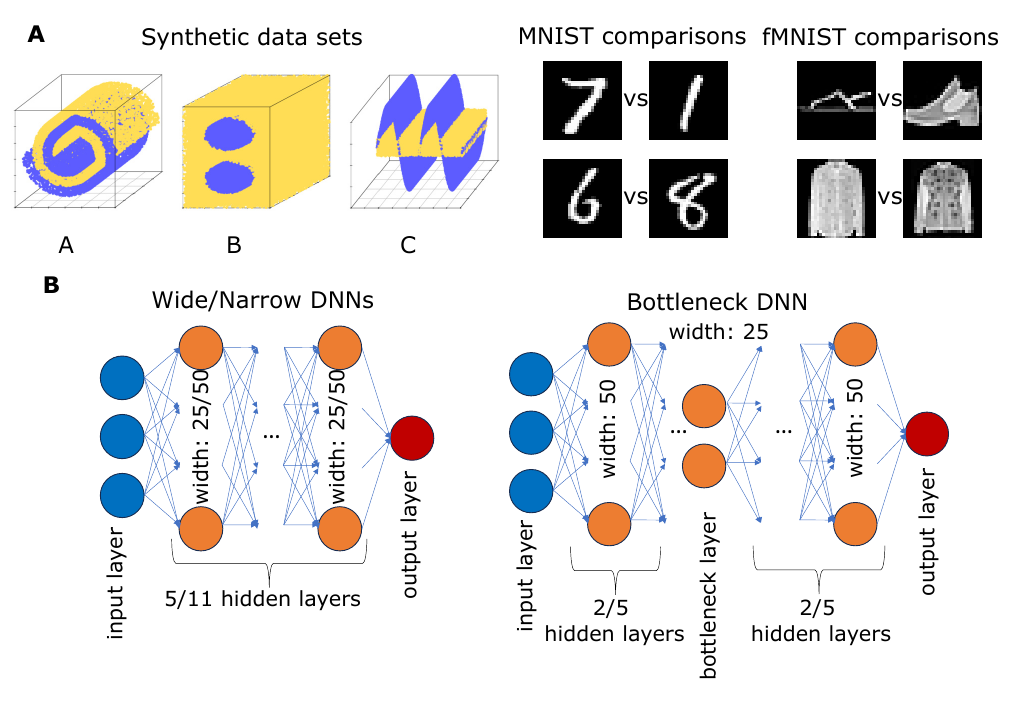}
\caption{{\it{Data sets for binary classification and DNN architectures trained.}} \textbf{A}. Three synthetic data sets A, B and C describe binary classification problems with different degrees of geometric and topological entanglement. We also considered two binary classification problems from the MNIST data set: distinguishing similar looking numbers (`1' vs `7' and `6' vs `8'). Finally, we considered two binary classification problems from the fashion MNIST data set: distinguishing similar looking items of clothing (`sandals' vs `ankle boots' and `shirts' vs `coats') \textbf{B}. For each problem, three different DNN widths were considered: narrow (25 nodes wide); wide (50 nodes wide); and bottleneck, as shown. For each choice of width, two depths were trained: shallow (5 hidden layers) and deep (11 hidden layers).}
\end{figure}

For each synthetic data set we trained multiple feed-forward DNNs on a training set of 1,000 data points, uniformly sampled from each data set A, B and C. For the MNIST problems, we used the full training data set containing 6,742 `$1$' digits, 6,275 `$7$' digits, 5,918 `$6$' digits and 5,851 `$8$' digits. For the fMNIST problems we used the full training set containing $6,000$ images each of `sandals', `ankle boots', `shirts' and `coats'. For each problem we trained ensembles of DNNs with three widths (Figure 2B): narrow (in which each layer has width 25), wide (in which each layer has width 50) and bottleneck (in which the first layer has width 50, compressing to a middle layer bottleneck of width 25 nodes and expanding back to 50 nodes on output). For each width we trained a shallow version (with 5 layers) and a deep version (with 11 layers). To account for stochastic effects, and provide a sufficient number of trained classifiers for subsequent statistical analysis, we trained 36 DNNs for each combination of width and depth. 

For each synthetic data set A, B and C we also extracted a test data set $X_{\textrm{test}}$ consisting of 1,000 data points disjoint from the respective training sets. For the MNIST problems, the test data contains 1,135 `$1$' digits, 1,028 `$7$' digits, 958 `$6$' digits and 974 `$8$' digits, disjoint from the training data. For the fMNIST problems, the test data contains 1,000 images of each item of clothing. Following previous studies \cite{Naitzat2020,Brahma2016} we investigated DNNs that are near-perfect classifiers: that is, $>99\%$ test accuracy for data sets A and B and MNIST and $>91\%$ test accuracy for data set C (where perfect accuracy is, by construction, impossible). For fMNIST no DNN trained reached $>99\%$ accuracy, so near-perfect classification was considered at $>98\%$ accuracy for the `sandals' versus `ankle boots' comparison and at $>90\%$ for `shirts' versus `coats'. In total, 151 of 1512 DNNs trained ($9.9\%$) failed to reach this level of test set accuracy, resulting in 1,361 near perfect classifiers ($\geq 25$ of each DNN architecture, for each data set).

%\subsection{The number of nearest neighbours $k$}
\subsection{Discretising the manifold}

The Ricci coefficient defined above depends upon the parameter $k$, the number of nearest neighbours in the graph $\mathcal{G}_{k}(X_{\textrm{test}}^{l})$, which captures the appropriate scale of the manifold and its representations by the DNN. It is not \emph{a prori} clear which choice of $k$ will be optimal to observe global Ricci network flow-like behaviour for a given data set and DNN, nor whether there always exists a $k$ that yields a negative Ricci coefficient. 

We assume that the optimal $k$, if it exists, depends on input data, DNN width and depth. In order for the Ricci coefficient to be defined, $k$ must be sufficiently large to ensure $\mathcal{G}_{k}(X_{\textrm{test}}^{l})$ consists of a single connected component for all $l$, otherwise there will exist $i,j \in V$ such that $\gamma_{k}^{l}(i,j)$ is not-defined. For each DNN and data set, we therefore considered a range of values for $k$ from the smallest $k$ that ensures that $\mathcal{G}_{k}(X_{test})$ is a single connected component, up to at least $\sim 20\%$ of the test data set size. For the synthetic data sets where the test data set comprised 1,000 points, we considered $k \in \{6,7,9,10,15,18,20,30,50,90,100,120,140,150,160,180,200 \}$. For MNIST and fMNIST where the test data set comprised $\sim 2,000$ points we considered $k \in \{ 10,20,30 \allowbreak 
 ,50,90,100,120,150,250,350,500 \}$.

For each DNN of a given width and depth trained on a given data set, we calculated $\eta_l$ and $\textrm{Ric}_l$ and aggregated values across the $\geq 25$ DNNs trained to near-perfect accuracy to compute a single aggregated Ricci coefficient for each considered value of $k$. This uncovered a complex relationship between the aggregated Ricci coefficient and $k$, which is influenced by data set and DNN width and depth (Figure 3). For the synthetic data sets, hierarchical clustering revealed the relationship between aggregated Ricci coefficient and $k$ to be strongly influenced by data set, while for the MNIST and fMNIST data sets, DNN depth was a more dominant factor.

A value of $k$ admitting a negative aggregated Ricci coefficient was found for every binary comparison, however the range of $k$ values varied significantly. For synthetic data sets the value of $k$ yielding the most negative aggregated Ricci coefficient was typically low (data set A: $k=3\%-10\%$ of $|X_{\textrm{test}}|$, data set B: $k=0.6\%-1\%$ of $|X_{\textrm{test}}|$, data set C: $k=0.6\%-5\%$ of $|X_{\textrm{test}}|$). Conversely for the MNIST and fMNIST data sets the $k$ yielding the most negative aggregated Ricci coefficient was much higher (MNIST `1' versus `7' $k=16\%-23\%$ of $|X_{\textrm{test}}|$, MNIST `6' versus `8' $k=18\%-25\%$ of $|X_{\textrm{test}}|$, fMNIST `sandals' versus `ankle boots' $k=12.5\%-25\%$ of $|X_{\textrm{test}}|$, fMNIST `coats' versus `shirts' $k=25\%$ of $|X_{\textrm{test}}|$).

Increasing $k$ tends to increase the average curvature of the network since a low value of $k$ results in a sparser, more tree like structure, while a higher value will increase the expected number of cliques. The finding that our synthetic data sets show stronger Ricci flow-like behaviour at low values of $k$, suggests that progressing through the layers of a well trained DNN may favour increasing distances between points from different classes. Conversely, for the MNIST and fMNIST data sets stronger Ricci flow-like behaviour occurs at higher values of $k$, suggesting that a well trained DNN may instead favour decreasing distances between points of the same class.

\begin{figure}
\centering
\includegraphics[width=1\textwidth]{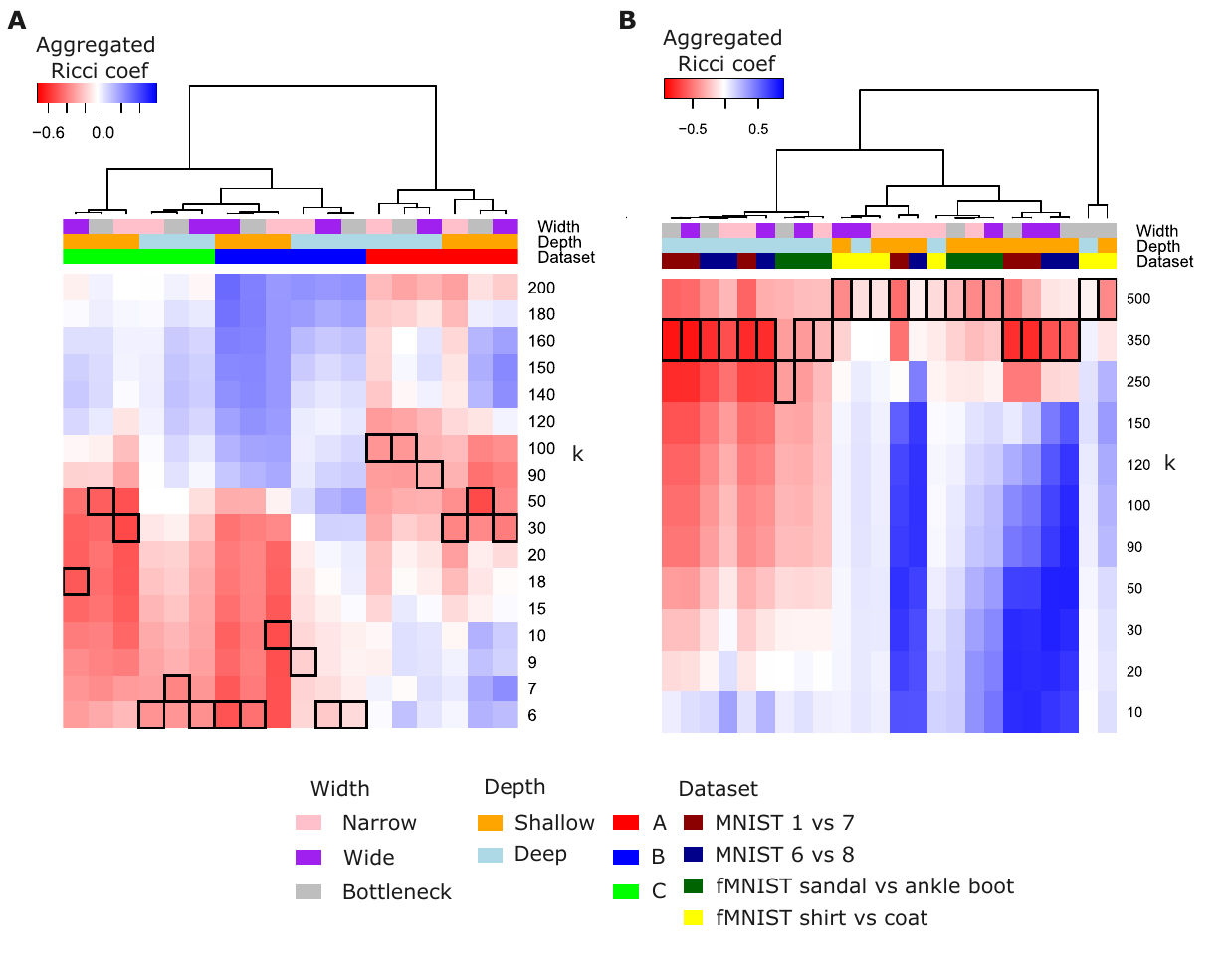}
\caption{{\it{Ricci flow-like behaviour and the number of nearest neighbours $k$.}} \textbf{A}. Heatmap of aggregated Ricci coefficients, computed across $\geq 25$ DNNs of a given width and depth trained on a given data set, for various values of $k$ evaluated for synthetic test data sets A, B and C of 1,000 points. \textbf{B}. Heatmap of aggregated Ricci coefficients, computed across $\geq 25$ DNNs of a given width and depth trained on a given data set, for various values of $k$ evaluated for binary comparisons in the MNIST and fMNIST data sets with test data sets of $\sim 2,000$ points. Black boxes outline the value of $k$ yielding the most negative aggregated Ricci coef. For both heatmaps the dendrogram shows the results of hierarchical clustering using the aggregated Ricci coefficients as a feature. We see that higher values of $k$ are required to observe Ricci flow-like behaviour in the MNIST data sets compared to the synthetic sets.}
\end{figure}

\subsection{Ricci flow-like behaviour indicates DNNs perform dataset-specific manifold deformations}

To measure global Ricci network flow-like behaviour in individual trained networks, we calculated the Ricci coefficients, given in Eq.~\ref{RC}, using the value of $k$ which yielded the most negative aggregated Ricci coefficient for the associated data set, DNN width and depth, established above. $\mathcal{G}_{k}(X_{\textrm{test}}^{l})$ did not consist of a single connected component, for at least one value of $l$, for this $k$ in 134/1361 ($9.8\%$) DNNs and these were excluded from further analysis. As expected by construction the majority of analysed DNNs (1144/1227, $93.2\%$) displayed negative Ricci coefficients. Representative plots of total curvature at layer $l$, against total change in distance between point pairs at layers $l+1$ and $l$, for each data set and various DNN architectures are shown in Figure 4. Ricci coefficients for each DNN alongside test set accuracy are provided in Table S1.

As a low optimal value of $k$ results in low curvature representations of the data, we posited that Ricci flow across layers of a well trained DNN under such a $k$ will prefer to separate points from distinct classes, rather than converge points from the same class. In line with this hypothesis, we found that for DNNs trained on synthetic data (where a lower $k$ is optimal), total curvature decreases as we progress through the layers. Conversely, for DNNs trained on MNIST and fMNIST (where a higher $k$ is optimal), curvature increases through the layers of DNNs, suggesting that DNNs trained on these data sets prefer to push similar points together (Figure 4). 

\begin{figure}
\centering
\includegraphics[width=1\textwidth]{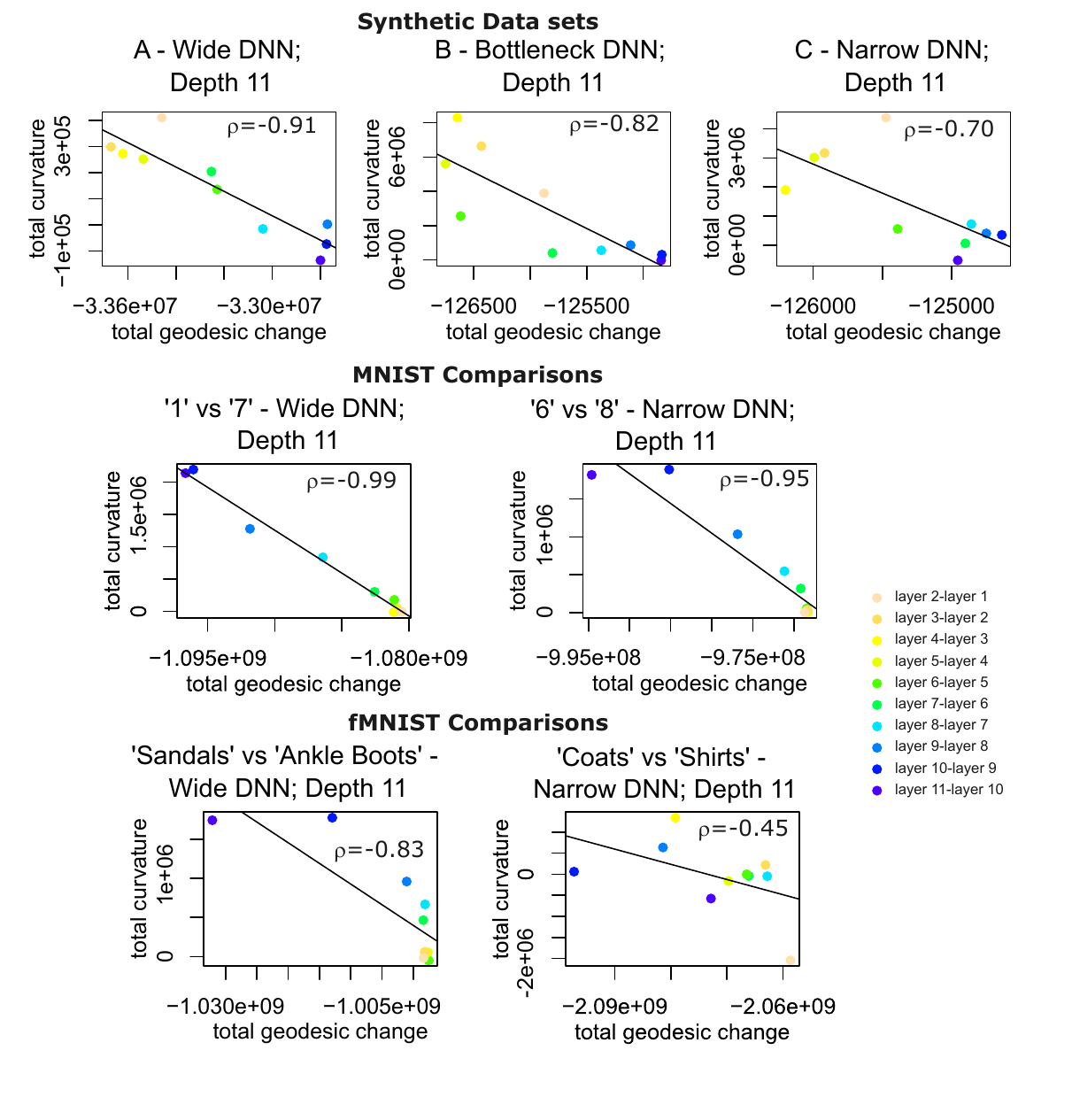}
\caption{{\it{Ricci flow-like behaviour has different implications for different data sets}} Scatter plots display total curvature at layer $l$ against total change in distance between point pairs at layers $l+1$ and $l$, for each data set and various DNN architectures. The Ricci coefficient ($\rho$) for each DNN is presented on each plot. For the synthetic data sets, total distance change between points increases through the layers and curvature drops, implying a separation of points from different classes. Conversely for MNIST and fMNIST total distance change decreases through the layers and curvature increases, implying an aggregation of points from the same class.}
\end{figure}

\subsection{Global Ricci network flow-like behaviour and test set accuracy}

Although we selected well trained DNNs with an accuracy $>90\%$, there was variation in the accuracy achieved above this threshold. To determine whether the Ricci coefficient associated with this variation in accuracy, we fit a multivariate linear model of test accuracy as a function of Ricci coefficient, data set, DNN width and DNN depth across the 1,227 DNNs trained to sufficient accuracy, on which a Ricci coefficient was calculable. We found that the Ricci coefficient negatively associated with test accuracy independently of data set or DNN (regression coefficient $t$-value$=-2.27$, $p=0.023$, Table S2), indicating that stronger global Ricci network flow-like behaviour associates with higher test accuracy.

\section{Discussion}

Central to the power of DNNs is their ability to generalise, which has been related to their capacity to identify broad geometric principles \cite{stephenson2021geometry}. Accordingly, a number of studies have shown that consistent changes occur in the discrete geometry and topology of data clouds as they pass through the layers of a well-trained DNN classifier \cite{Brahma2016,Naitzat2020,cohen2020separability}. Here, we have highlighted a conceptual similarity between such behaviour and that expected by Ricci flow. By adapting recently developed tools to study discrete Ricci flow \cite{Ni2019,weber2016,baptista2023charting}, we constructed a framework to assess the significance and strength of global Ricci network flow-like behaviour in data clouds passing through DNN layers. In particular, we have introduced an easily computable metric that quantifies Ricci flow-like behaviour, that we refer to as the Ricci coefficient. Using this metric, we provided a framework to identify a scale at which Ricci flow-like behaviour can be observed in a set of DNNs with a given architecture trained on a given data set. We found that this scale can provide insight into how binary classification is performed for a given data set by DNNs, in particular revealing the trade off between separation of disparate points and convergence of similar points. Our framework is computationally and conceptually simple, yet yields meaningful results relating DNN accuracy to Ricci flow-like behaviour. This motivates further investigation and deeper theoretical characterisation of how we can use Ricci flow-like behaviour to facilitate transparency in deep learning.

There is currently no consensus on how to best optimise neural network architectures. Common general approaches such as cross validation are typically computationally expensive, precluding their use in some applications \cite{rao2008dangers}. There is, therefore, a need for design processes that leverage theoretical understanding of deep learning \cite{calvo2023optimal}. Here, we have found that the Ricci coefficient of a trained DNN computed at an appropriate scale is positively associated with classifier test accuracy, independently of DNN architecture or data set, suggesting that it may be a useful model selection tool. In particular, our numerical experiments suggest that a more negative Ricci coefficient is indicative of a DNN architecture that is appropriately matched to the problem at hand, and able to generalise well; while a less negative Ricci coefficient is indicative of an architecture that is over-fit to the problem at hand an unable to generalise well. While preliminary, this observation suggests that further consideration of metrics based on Ricci flow and associated geometric tools may aid in the design, development and optimisation of neural network architectures. 

To conclude, we note that there is a distinction between global and local Ricci flow in our framework. Here, we have only considered the global geometric changes in data as it passes through a neural network, as summarised in a single constant — the Ricci coefficient. We find that for a suitable choice of scale, changes in total distance between data point pairs are typically negatively correlated with an assessment of the total curvature between pairs of connected data points. This finding suggests that when the total curvature of a data set representation at a given layer is negative, the total distance between data point pairs at the next layer will increase, which is the behaviour expected from a global Ricci network flow. Under a local Ricci flow, a stronger condition pertains: the geodesic distance between \emph{any} pair of data points will be inversely correlated to the curvature along the shortest path connecting those specific points. While a local Ricci flow implies a global flow, the converse is not true. Indeed, intuitively, we do not expect that deep learning is associated with local Ricci flow for two reasons. First, this would imply that trained DNNs could be specified by a single parameter (the strength of the local flow), a highly unlikely outcome, except for perfect classifiers. Second, for data sets describing entangled classes an element of `unraveling' will likely be required before distinct data class manifolds are separable, as has been observed in previous studies of topological simplifications in deep learning \cite{ansuini2019intrinsic,magai2022topology}. During this unraveling phase, strongly connected data points may move away from one another — the opposite behaviour to that expected under a local Ricci flow. Our finding that global Ricci network flow-like behaviour associates with greater classification accuracy does, however, suggest that distances between many (but not all) data point pairs may indeed change according to a local Ricci flow, and, moreover, the proportion of data point pairs for which local Ricci flow pertains is positively associated with classifier accuracy. We hypothesise that data point pairs that exhibit strong local Ricci flow-like behaviour under the action of a DNN are precisely those points that are well characterised by the DNN, while data point pairs that do not exhibit local Ricci flow-like behaviour are those that the DNN struggles to distinguish. Further investigation of local and global Ricci network flow dynamics in DNNs could thus provide a novel approach to point specific uncertainty quantification, fairness assessment, and detection of anomalies and out of distribution data.

\section{Supplementary data}

{\bf{Table S1: }}{\it{Summary of DNNs trained and Ricci flow behaviour for each data set.}} The table summarises the following variables for each of the 1,227 DNNs trained to high accuracy on which Ricci coefficients were computable: adjusted Ricci coefficient, test set accuracy, DNN width, DNN depth and data set the DNN was trained on.\\

{\bf{Table S2: }}{{\it{Ricci flow-like behaviour associates with test set accuracy, independently of DNN width, depth and data set}}. \rm{The table summaries the results of multivariate linear regression modelling of test set accuracy as a function of adjusted Ricci coefficient, data set, DNN width and depth across the 1,227 DNNs trained to high accuracy for which Ricci coefficients were computable. We observe a significant negative association between  Ricci coefficient and test set accuracy, independently of other variables.}}\\

\section*{Code availability}
The codes to run the Ricci flow-like analysis are available on GitHub at the following link: \href{https://github.com/anthbapt/Ricci-NN}{https://github.com/anthbapt/Ricci-NN}. Both Python and R versions of the code are provided, along with associated data from the Fashion-MNIST dataset.

\section*{Competing interests statement}
The authors declare no conflict of interest.

\section*{Acknowledgments}

All authors gratefully acknowledge support from the Turing-Roche strategic partnership.

\bibliography{bibliography}
\bibliographystyle{unsrt}

\newpage
\end{document}